\documentclass{article}
\usepackage[utf8]{inputenc}
\usepackage[style=apa,backend=biber]{biblatex}
\usepackage{graphicx}
\usepackage{eurosym}

\title{\huge{Neuromorphic Computing - An Overview}}
\author{Benedikt Jung\textsuperscript{*}, Maximilian Kalcher\textsuperscript{*}, Merlin Marinova\textsuperscript{*}, \\Piper Powell\textsuperscript{*}, Esma Sakalli\textsuperscript{*}\\[0.5em]\small \emph{\textsuperscript{*}Osnabrück University}}

\date{}
\begin{document}

\maketitle

\begin{abstract}
    With traditional computing technologies reaching their limit, a new field has emerged seeking to follow the example of the human brain into a new era – neuromorphic computing. This paper provides an introduction to neuromorphic computing, why this and other new computing systems are needed, and what technologies currently exist in the neuromorphic field. It begins with a general introduction into the history of traditional computing and its present problems, and then proceeds to a general overview of neuromorphic systems. It subsequently discusses the main technologies currently in development. For completeness, the paper first discusses neuromorphic-\emph{style} computing on traditional hardware, and then discusses the two top branches of specialized hardware in this field – neuromorphic chips and photonic systems. Both branches are explained as well as their relative benefits and drawbacks. The paper concludes with a summary and an outlook on the future. 
\end{abstract}
\bigskip
\begin{center}
{\footnotesize{This article was originally published on April 27, 2023, in \linebreak \textit{Cognitive Science Student Journal, 2023, Volume 7, pp. 1–10.}}}
\end{center}

\section{Introduction}

Modern computing has a problem. Since the mid-1900s, computing has been carried out primarily on machines following the so-called “von-Neumann" architecture (here abbreviated VNA), named after John von Neumann and dating all the way back to his work on the ENIAC project during the Second World War. That architecture, and the fundamental architecture of all standard computers since, consists of a memory storage component (in which both the computer’s data and its operating instructions are stored), a central processing unit (CPU) connected to that memory via a digital bus, and input and output components \parencite{ENIAC}. Over 60 years later, we are now seeing the limits of how far that architecture can take us.\\

From its start, a key issue with VNA was the digital bus connecting its memory and processing components. The processing component needs to operate on the data contained in the memory, but the intervening digital bus limits the amount of data that can pass from the memory to the processor at one time, as well as the speed of that passing. This critical “bottleneck” in the system has only become more of a concern as the average speed of CPUs has increased and memory storage has grown. Modern CPUs working on modern tasks require large amounts of data to be made available at rapid speeds, placing an increasing demand on the bottlenecking digital bus separating that data from the processor that needs it. This problem in itself will only continue to grow worse, and it is not the only problem being faced \parencite{shastri2017}.\\

The number of transistors on a single microchip doubles approximately every two years, according to the Moore’s Law principle \parencite{andreoli}, and average clock speeds and power efficiency have doubled at roughly the same rate since the mid-1900s \parencite{shastri2017}. This trend has led to enormous advances in modern computing, but we are finally hitting a plateau where traditional architecture simply cannot be advanced any further \parencite{marr,christensen} – a major concern for the world at large, as our data and processing demands will only continue to increase.\\

Standard VNA machines also differ in their construction to the human brain, which may limit their ability to achieve human-like intelligence \parencite{padhyegurjar}. At the present time, the best supercomputers in the world (based on the traditional architecture) are 8 orders of magnitude less computationally efficient than the human brain, maxing out at 100pJ/MAC (multiply accumulate operations). The human brain by comparison has a computational efficiency of less than 1 aJ/MAC, and achieves an incredible speed of 10\textsuperscript{18}MAC/s with 20W of power - the same amount required to turn on the average light bulb \parencite{shastri2017}. \\

In order to solve these problems, a new generation of computing systems has entered the stage, aiming to boost efficiency and capacity by harnessing the power of biologically-inspired architectures and algorithms – neuromorphic computing. This growing field aims to follow the brain's example towards a new era of computational power and efficiency. We will provide here an introduction to this exciting avenue and the main technologies currently in development.

\section{Introduction to Spiking Systems} 

Neuromorphic engineering aims to create computing hardware that mimics the nervous system of the human brain. Hence, these hardware systems are based on the structures, processes, and capacities of neurons and synapses in the brain. Neuromorphic hardware operates in a spiking paradigm where a single unit, analogous to a single neuron in the brain, is only active when it receives or emits information in the form of electric impulses called "spikes," meaning that these systems operate purely on sparse binary signals in a purely event-driven manner \parencite{roy}. In contrast, conventional computing systems, regardless of the input, have all units active at all times, which results in unnecessary time and energy consumption.\\ 

The implementation of neuronal and synaptic computations through spike-driven communication enables energy-efficient and more precise machine learning since spiking systems use time as an additional input dimension. Contrary to conventional computing, such as computers based on VNA, neuromorphic systems consist of non-volatile memory and analog processing circuits and can thus store and process more digital information while consuming less power, a key advantage in a world with an ever-increasing demand for efficient and mass-storage data processing capacities \parencite{christensen}.\\

In the following section, we will cover the use of spiking systems on two of the top neuromorphic platforms currently used - neuromorphic chips and photonic systems. For completeness, we first begin with a review of spike computing on conventional hardware. \\

\section{Technologies}

\subsection{Neuromorphic Computing on Conventional GPUs}

Conventional GPUs, or graphical processing units, are a type of processor that is commonly used in computers to handle graphical data and perform complex calculations. They were originally designed for gaming and other graphical applications, but have recently been used in a variety of other fields, including machine learning and artificial intelligence. Unlike traditional central processing units (CPUs), which are designed for general-purpose computing tasks, GPUs are specifically optimized for parallel processing and are able to process thousands of small independent tasks simultaneously. This allows them to deliver significant performance gains over CPUs in graphics-intensive tasks, making them an essential component of modern computing systems.\\

Current advances in many domains where deep neural networks (DNNs) are applied have shown that more computing power generally accounts for half or more improvement in outcomes. This is not only due to the computing resources themselves, but also the algorithms that implicitly change in order to harness these resources effectively \parencite{thompson}. Even for Spiking Neural Networks (SNNs), which operate on the newer spiking principle outlined above, GPUs have been the primary computing resource in many applications, thanks to their availability and adaptability. GPUs are particularly useful in SNNs because they can offer some parallelism and high performance, which is not present in traditional neural networks, though at the cost of higher power consumption. This allows for faster and more efficient processing of data, leading to better accuracy and faster inferencing times \parencite{huyhn}.\\

Moore's Law has been a key principle reflected in the rapid advancement of conventional GPUs, which are used in a wide range of applications from gaming and entertainment to scientific research and machine learning. Modern conventional chips have transistors that are now as small as 5nm, which is smaller than most viruses. Going a step further into atom-size territory would make things not only extremely expensive but also challenging. In these dimensions, the limits of physics are reached and the switching of transistors could be influenced by random statistical fluctuations \parencite{andreoli}, the reason why traditional computing, even with GPUs, is reaching its limits.\\

The most commonly used GPU in large-scale deep learning settings, and widely used for cloud computing in servers, is the Nvidia Ampere A100. It computes roughly 19.5 TFLOPS (tera floating-point operations per second) on double-precision floating-point operations, making it one of the most powerful GPUs available. This chip is popular due to its efficiency when handling large sparse operations, alongside its 40 GB of on-chip HBM2 memory and more than 2TB/s of memory bandwidth. Besides supporting the most common deep learning frameworks such as TensorFlow, PyTorch, and others, it can also be used for general purpose computing, like any other GPU. However, the high performance of the A100 means a larger power draw, with this GPU drawing up to 400W of power consumption on maximal workloads \parencite{svedin}. The pricing is also likely to be a barrier for the regular consumer, with the A100 costing 11.599,00\euro \  MSRP. \\

\begin{figure}
    \centering
    \includegraphics[width=0.65\textwidth]{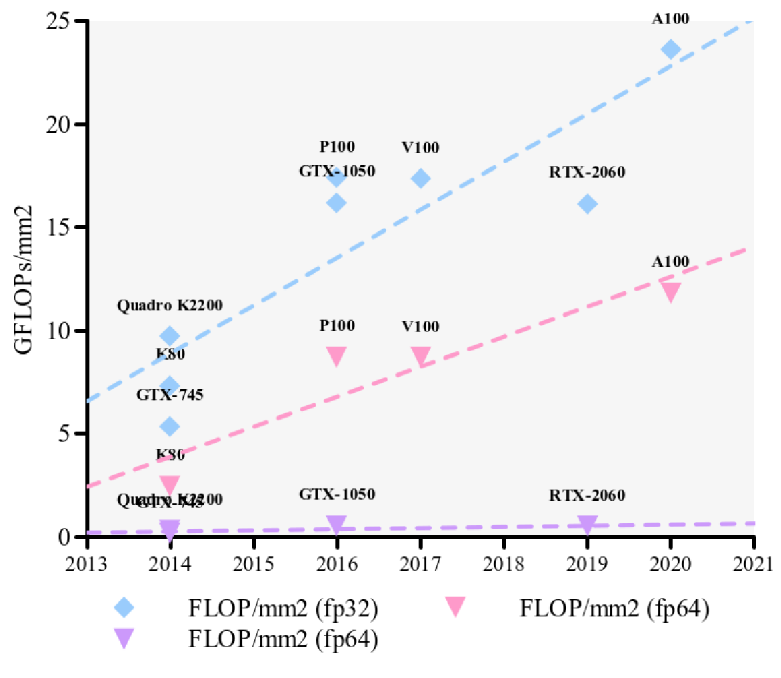}
    \caption{The compute density of modern GPUs. Taken from \parencite{svedin} in compliance with the figure's CC BY-NC-ND license allowing attributed free use.}
\end{figure}

\subsection{Neuromorphic Chips}
Conventional hardware, based on the VNA, separates memory and processing, increasing power consumption and making it difficult to accurately simulate the brain's highly interconnected network of neurons, especially when using architectures such as Spiking Neural Networks (SNNs) \parencite{ivanov}. To mimic the way the brain works on the software side, you also need hardware with underlying mechanisms that are also fundamentally aligned with the brain. \\

One such hardware approach is the neuromorphic chip. In practice, neuromorphic chips are used to create artificial intelligence systems that can interact with the environment in a more natural way. For example, they can be used to develop robots that can navigate complex environments and respond to changing conditions in real time, or to create intelligent systems that can understand and interpret sensory data from the world around them.\\

Neuromorphic chips are designed to mimic the structure and function of the human brain, using a network of artificial neurons and synapses, and they process based on the massively parallel, event-driven, and analog properties of the brain \parencite{boahen}: 

\begin{itemize}
    \item[$-$] Analog Computation: Neuromorphic systems use analog computation, which is a type of computation that uses continuous, physical quantities to represent data and perform calculations. This is in contrast to digital computers, which use discrete, binary digits (0s and 1s) to represent data and perform calculations. Analog computation allows neuromorphic systems to process information more efficiently and more accurately than digital computers.
    \item[$-$] Event-Driven Computation: Neuromorphic systems are event-driven, which means that they only perform calculations when necessary rather than continuously like traditional computers. This allows them to conserve energy and reduce power consumption, which is a key advantage over conventional systems.
    \item[$-$] Mixed-signal Integration: Neuromorphic systems use both analog and digital signals to represent and process information, which allows them to combine the strengths of both types of signals. For example, analog signals can be used to represent continuous, physical quantities more accurately, while digital signals can be used to perform logical operations more efficiently.
\end{itemize}

One attempt to make this computing architecture conventional is the neuromorphic chip Loihi by Intel. Loihi, being specialized for specific SNNs, uses a network of physical artificial neurons and synapses, which are connected in a manner that is similar to the way neurons are connected in the human brain. This allows it to process information in a more brain-like way, using techniques such as event-driven computation and mixed-signal integration. It is also highly energy-efficient, and can be used for real-time, low-power applications. The newest generation of this chip (Loihi 2), already includes up to 1 million integrated neurons, with 10x faster processing capability, 60x more inter-chip bandwidth and 15x greater resource density than the first generation. The chip itself is currently not publicly available for purchase \parencite{loihi}.\\ 

\subsection{Photonic Systems}

Traditional electronic systems face numerous limitations, including limited bandwidth and speed, limited communication distance, and crosstalk across channels. In contrast, systems based on light can operate at hyper-fast speeds, are more efficient, have a large bandwidth, can communicate over considerable distances, and exhibit low levels of crosstalk while simultaneously being capable of sending multiple communication streams over the same channel \parencite{robertson,owen,de2017progress}. Neuromorphic photonics takes advantage of these properties and aims for brain-like computations and functionalities with a photonic platform \parencite{shastri2017}, potentially far outstripping even the performances of other benchmark neuromorphic systems (see Figure 2).\\

\begin{figure}[h!]
    \centering
    \includegraphics[width=0.9\textwidth]{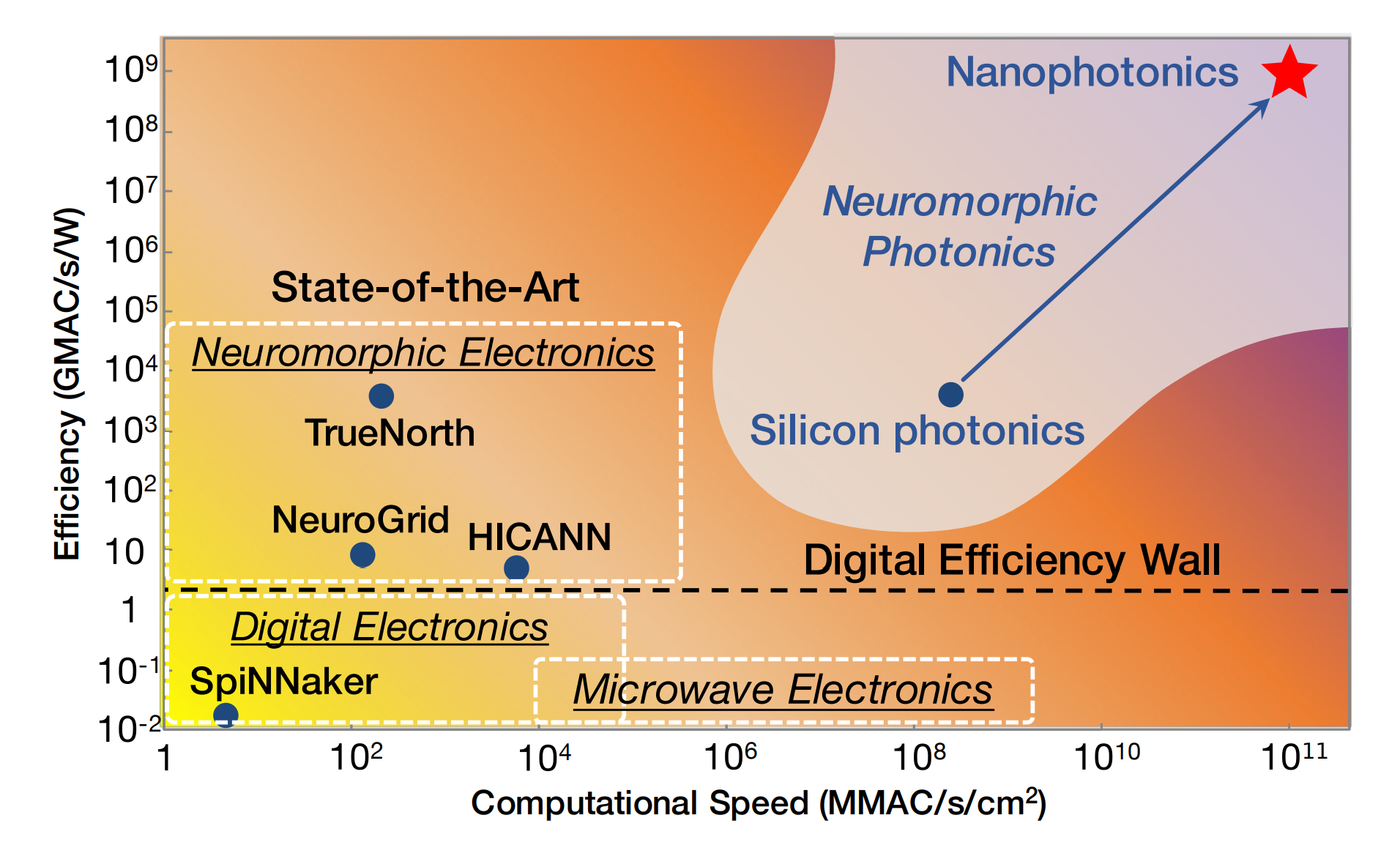}
    \caption{Theoretical speed versus efficiency of neuromorphic photonic systems as compared to other neuromorphic systems. Taken from \cite{shastri2017} in compliance with the figure's CC BY-NC-ND license allowing attributed free use.}
\end{figure}

One key advantage of photonic neuromorphic hardware is that it does not rely on new technologies being developed. It instead is a repurposing of existing hardware already used in the telecommunications field (typically hardware designed for the 1310 and 1550nm wavelengths). In terms of sheer speed, these systems can clock out at 9 orders of magnitude faster than biological neurons and 6 orders of magnitude faster than electronic artificial neurons, depending on the exact medium used. Since the field’s start in 2010, it has grown to include various technologies based in various photonic media, including crystal structures, semiconductor amplifier modulators, and laser-based systems \parencite{robertson}.\\ 

Another main benefit of using photonic systems in neuromorphic computing is their high speed and parallel processing capabilities. Optical systems can operate at very high speeds, on the order of up to 10 GHz and more \parencite{brunner2016all}, which is much faster than the speeds achievable by electronic systems. This high speed is due to the fact that light travels much faster than electrons, and can be used to transmit and process information in parallel across multiple channels \parencite{de2017progress}. This makes optical systems well suited for tasks such as real-time image and video processing, which require fast and efficient processing of large amounts of data.\\

In addition to their high speed and low power consumption, optical systems have other unique features that make them attractive for neuromorphic computing. For example, optical systems can exhibit non-linear behavior \parencite{rasmussen2020all}, which is similar to the non-linear behavior of neurons in the brain \parencite{mckenna1994brain}. This non-linear behavior can enable optical systems to perform complex computations, such as pattern recognition and classification, that are difficult to achieve with linear systems \parencite{cai2022broadband}.\\

As laser systems are the most common form of photonic neuromorphic system \parencite{robertson}, we cover them in more depth here, before briefly mentioning other technologies in this area.  

\subsubsection{Lasers}
Many laser-based neuromorphic photonic systems (LNPSs) are based on semiconductor lasers because these can exhibit biologically valid behavior such as excitability and non-linear dynamics off-the-shelf. Of particular interest within this family of lasers are the vertical-cavity surface-emitting laser diodes (VCSELs), which are compact, power-efficient, low cost, and can operate at high speeds at the standard telecommunications wavelengths. VCSELs are also an established technology and already implemented in everything from automotive sensors to bar code scanners, greatly reducing the amount of up-front development required to implement an LNPS with this type of laser diode as its foundational unit. In terms of brain-like computing, VCSELs are also capable, without modification, of exhibiting and supporting biologically valid behaviors such as threshold-and-fire operation, tonic and phasic spiking, and spike rate-encoding, all while operating at sub-nanosecond speeds \parencite{robertson,owen}. As they are a prime choice in the field, we will introduce the concept of LNPSs with a foundation in VCSELs here.\\

In LNPSs, individual VCSELs act as the individual neurons in the system and react to light either from an external light source or from other VCSELs. A different device, such as an optical circulator or intensity modulator, is then placed at the output point of a given architecture and converts the final output of the lasers to an output usable in tasks ranging from visual data pre-processing \parencite{robertson} to classification tasks \parencite{owen}. We will first discuss how signals are transmitted through VCSELs and then briefly discuss how these diodes can support network tasks and modelling.\\

Within a LNPS, each VCSEL reacts to changes in the incoming light it receives as input. In its resting state, a single VCSEL is “locked” to that light source if it receives a steady stream of light from it with no interruptions. Introducing interruptions to that light source causes the VCSEL to “unlock” from it, a behavior analogous to a spike-event in a biological neuron. These events can occur on an extremely fast timescale, in the space of 100ps in comparison to the millisecond timescale of biological neurons. Modifying this spiking behavior involves adjusting the length of the light interruption, with longer periods of interruption resulting in increased spike activity (to understand this, it is helpful to remember that the stimulus in this case is the \emph{absence} of the light). A VCSEL neuron can also be set up to act in an inhibitory fashion, emitting fewer spikes when exposed to longer light interruptions. Because this switching behavior is based on the turning off and on of the light source, spikes can occur as close as 500ps apart from each other, an extremely fast timescale \parencite{robertson}.\\

Linking multiple VCSEL neurons together (so that later neurons receive the output of other neurons as their input) allows these diodes to form a neural-style network. Memory can even be introduced into the system in the form of a delayed looping spike train which is fed back into a solitary VCSEL at regular intervals (in a one-neuron system) or is incorporated into the connection between two VCSELs (in a multi-neuron system). When VCSELs are incorporated into a reservoir-style LNPS where the internal neuron connections are fixed, only the output weights for the network must be trained for a given task (via a simple matrix computation incorporating the target output and the actual output of the network), resulting in a vastly increased training efficiency. This specific architecture has proven successful in image classification tasks \parencite{owen}, and general networks of VCSELs have proven successful as the basis for LNPS models of biological systems such as bipolar and retinal ganglion cell circuits in the visual system \parencite{robertson}.\\

LNPSs have the potential to operate far faster and more efficiently even as compared to other neuromorphic technologies, and can far outstrip traditional computing. Because they can be based in existing technologies, LNPS are also highly usable and, as hardware like VCSELs are low-cost, have an advantage in terms of expense. In terms of accessibility, however, no LNPS is currently available for general purchase.\\

\subsubsection{Other Photonic Systems}

Spiking systems can be based on a number of other photonic technologies, such as silicon photonics, nanophotonics or metamaterials, which are artificially engineered materials that have properties not found in naturally occurring materials \parencite{shastri2021photonics,sylvestre2021neuromorphic}. Researchers at the Massachusetts Institute of Technology have also developed a neuromorphic computing system based on optically-controlled phase change materials, which uses laser pulses to control the phase of the materials, which can be used to store and process information in a manner similar to the way neurons in the brain process information \parencite{han2017optically}.\\

\section{Conclusion}

Computing with VNAs has advanced considerably over the years, but it is finally reaching its limit, necessitating that we look for new solutions for our future computing needs. A highly promising solution is neuromorphic hardware, which offers efficient and high-capacity computing power by harnessing the advantages of brain-like architectures and processes. At present, there is no one system which is certain to be the neuromorphic platform of the future, but neuromorphic chips and photonic systems are both key contenders.\\

Both systems are currently in development and both have benefits and drawbacks. Chips are highly efficient and already proving capable in numerous applications, but their physical basis limits their speed and scalability. Photonic systems, and especially laser systems, are also extremely power-efficient and can operate on even smaller timescales and over larger distances, but do suffer from issues ranging from limited wavelengths available to the hyper-precise and challenging component calibration required to operate them \parencite{wan2022neuromorphic}. \\

At this time, neither technology is publicly available or integrated into widespread commercial devices or applications, but either technology could easily become the future backbone of a new neuromorphic computing era. And while current large-scale operations still rely heavily on the parallel software-coupled and traditional VNAs such as the A100, the potential power of neuromorphic hardware is too great to ignore, though what exactly those systems might look like in the future is unknown. Regardless of which neuromorphic platform ultimately gains dominance, however, it is clear that the computing revolution behind their development is already upon us. 

\clearpage

\printbibliography
\end{document}